\documentclass{article}
\usepackage{spconf,amsmath,graphicx,hyperref}
\usepackage{amssymb}
\usepackage{cite}
\usepackage{balance}

\title{Reducing the modality gap enables multimodal embedding compression}

\title{From Gap to Compression: Compact Multimodal Representations via Contrastive Learning}
\title{Compressed Multimodal Representations via Contrastive Learning}
\title{Semantic Compression via Multimodal Representation Learning}
\name{Eleonora Grassucci, Giordano Cicchetti, Aurelio Uncini, Danilo Comminiello\thanks{This work was partially supported by the European Union under the Italian National Recovery and Resilience Plan (NRRP) of NextGenerationEU, Mission 4, Component 2, Investment 1.3, partnership on “Telecommunications of the Future” (PE00000001 - program “RESTART”) and partially by the \textit{Progetti di Ateneo} of Sapienza University of Rome under grant RM123188F75F8072 and RM1241910FC4BEEA.}}
\address{Dept. of Information Engineering, Electronics, and Telecomm., Sapienza University of Rome, Italy}
\begin{document}
%
\maketitle
\ninept
\begin{abstract}
Multimodal representation learning produces high-dimensional embeddings that align diverse modalities in a shared latent space. While this enables strong generalization, it also introduces scalability challenges, both in terms of storage and downstream processing. A key open problem is how to achieve semantic compression, reducing the memory footprint of multimodal embeddings while preserving their ability to represent shared semantic content across modalities.
In this paper, we prove a strong connection between reducing the modality gap, which is the residual separation of embeddings from different modalities, and the feasibility of post-training semantic compression. When the gap is sufficiently reduced, embeddings from different modalities but expressing the same semantics share a common portion of the space. Therefore, their centroid is a faithful representation of such a semantic concept. This enables replacing multiple embeddings with a single centroid, yielding significant memory savings. 
We propose a novel approach for semantic compression grounded on the latter intuition, operating directly on pretrained encoders. We demonstrate its effectiveness across diverse large-scale multimodal downstream tasks. Our results highlight that modality alignment is a key enabler for semantic compression, showing that the proposed approach achieves significant compression without sacrificing performance.
\end{abstract}
\begin{keywords}
Semantic Compression, Multimodal Learning, Modality Gap, Contrastive Learning, Embedding Compression
\end{keywords}
\section{Introduction}
\label{sec:intro}

Multimodal models have become a central tool in modern machine learning. By integrating data from different sources such as images, text, or audio, multimodal models are able to capture richer semantics than unimodal counterparts and are now used in a wide range of applications, from image classification and captioning to speech–vision systems \cite{Yoon2023HEARHE, Wang2024InternVideo2SV}. Large-scale frameworks such as CLIP \cite{Radford2021LearningTV} have shown that contrastive pretraining on paired data can align heterogeneous modalities into a shared latent space, providing representations that are transferable and effective across tasks \cite{CLAP2022, Girdhar2023ImageBindOE, cicchetti2025gram, zhang2023diagnosingrectifyingvisionmodels}. This has established contrastive learning as the \textit{de facto} paradigm for multimodal representation learning.

\begin{figure}
    \centering
    \includegraphics[width=\linewidth]{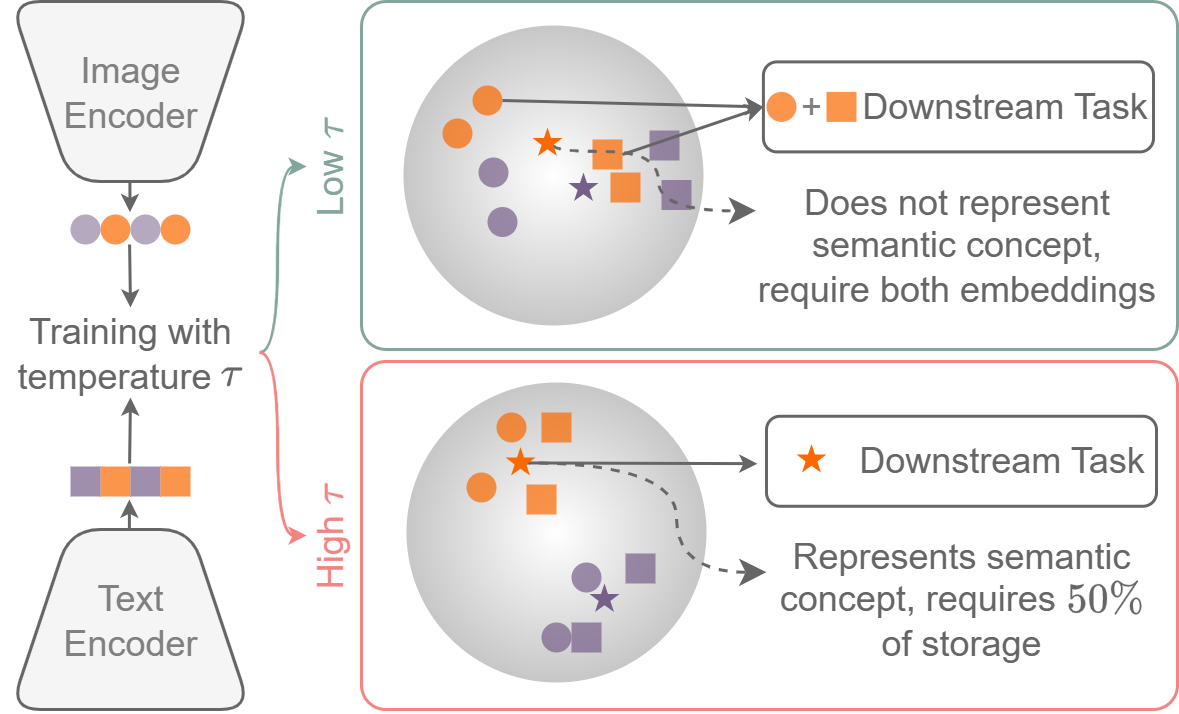}
    \caption{In the case of low temperatures, the modality gap prevents building a semantically-meaningful latent space and both the modality embeddings have to be used to perform downstream tasks. When the temperature is sufficiently high, the modality gap is reduced and the centroid becomes representative of a given class, thus it can be directly stored and used for downstream tasks, saving consistent storage without affecting performance.}
    \label{fig:overview}
\end{figure}

At the same time, it has been observed that multimodal embeddings trained with contrastive losses do not perfectly align \cite{liang2022mind}. Instead, embeddings often remain clustered by modality, a phenomenon referred to as the modality gap \cite{liang2022mind, iclr2024two, Eslami2024MitigateTG, Grassucci2025ClosingMedical}. In practice, this means that samples representing the same semantic concept may not lie on a single compact cluster but rather form parallel structures in the latent space.
In the literature, there is no clear pattern on whether the modality gap is harmful for downstream tasks or not. Some works claim that reducing the gap improves retrieval performance \cite{yaras2024explainingmitigatingmodalitygap}, others state that the modality gap has a slight correlation with downstream performance \cite{iclr2024two}. Recent works highlighted that the gap can be controlled through the temperature parameter in the InfoNCE loss \cite{Oord2018RepresentationLW}. Indeed, learnable or smaller temperature tends to preserve the gap, while fixed higher temperature reduces the modality gap \cite{yaras2024explainingmitigatingmodalitygap}.

Another critical issue concerns the efficiency of these models. Embeddings produced by multimodal encoders are typically high-dimensional, often 1024 or 2048 dimensions. Compressing such embeddings while preserving their semantic content is crucial for communication \cite{shen2025compressionpixelssemanticcompression, grassucci2025closing, Tian2025TWC, tong2024multimodalsemanticcommunicationgenerative} and efficiency in real-world downstream tasks \cite{Chen2019DifferentiablePQ, Li2024Survey}. Previous work on compression has focused primarily on unimodal settings \cite{Ding2025icassp}. For example, Dinu et al. \cite{dinu2025effective} studied the impact of temperature on the intrinsic dimensionality of text embeddings, showing that higher temperatures lead to lower intrinsic dimensionality and thus better retention after post-training compression. Matryoshka Representation Learning \cite{kusupati2022matryoshka} explicitly trains embeddings at multiple granularities, making them adaptable to various constraints. However, while these approaches demonstrate the feasibility of compressing unimodal embeddings, little attention has been devoted to compression in multimodal spaces, where challenges such as the modality gap are unique.


In this paper, we prove that effective semantic compression of multimodal embeddings can be enabled by reducing the modality gap. Our novel semantic compression approach relies on the key observation that when the gap is reduced through training with sufficiently high temperatures, the embeddings of different modalities for the same class become tightly aligned.
This allows us to represent each semantic concept with a single centroid embedding by averaging all instances across modalities, a straightforward yet effective compression method.
Figure~\ref{fig:overview} illustrates our intuition: when embeddings are trained at low temperatures, centroids are not representative due to the persistent gap. On the contrary, with higher temperatures, embeddings with the same semantics share a common centroid, which can be compressed without sacrificing performance.
Our main contributions can be summarized as follows:

\begin{itemize}
    \item We prove the connection between training temperature, modality alignment, and semantic compression for multimodal embeddings.
    \item We propose a novel semantic compression approach based on multimodal centroids that requires storing only one embedding per semantic concept, thereby eliminating redundancy across modalities.
    \item We demonstrate that these centroids can be further compressed with simple post-training methods such as random feature selection while retaining strong performance on downstream tasks.
\end{itemize}

\section{Proposed Method}
\label{sec:method}

Let us consider a batch $\mathcal{B} = \{(\mathbf{x}_i^1, \mathbf{x}_i^2, \dots, \mathbf{x}_i^M)\}_{i=1}^N$, with each sample observed in $M$ modalities. For each modality $m$, an encoder $f^m: \mathcal{X}^m \rightarrow \mathbb{R}^D$ maps the input $\mathbf{x}_i^m$ to a $D$-dimensional embedding vector $\mathbf{z}_i^m = f^m(\mathbf{x}_i^m)$, that is then normalized to a unitary norm vector.

\subsection{Effect of Temperature and Modality Gap}
A standard approach for learning multimodal embeddings is contrastive learning, 
where representations of matching pairs are pulled together while non-matching pairs are pushed apart. The most common formulation is the InfoNCE loss 
\cite{Oord2018RepresentationLW}, defined for a pair of modalities $m$ and $n$ as

\begin{equation}
\label{eq:infonce}
\mathcal{L}_{\text{InfoNCE}}^{m \to n} = - \frac{1}{N} 
\sum_{i=1}^{N} 
\log \frac{\exp(\text{sim}(\mathbf{z}_i^m, \mathbf{z}_i^n)/\tau)}{\sum_{j=1}^{N} \exp(\text{sim}(\mathbf{z}_i^m, \mathbf{z}_j^n)/\tau)},
\end{equation}

\noindent where $\mathbf{z}_i^m$ and $\mathbf{z}_i^n$ are the normalized embeddings of the $i$-th positive pair, $\text{sim}(\cdot,\cdot)$ denotes cosine similarity, $N$ is the batch size, and $\tau > 0$ is the temperature parameter. The complete loss is typically computed in both directions, i.e., $\mathcal{L}_{\text{InfoNCE}}^{m \to n} + \mathcal{L}_{\text{InfoNCE}}^{n \to m}$ and averaged.

\begin{figure}
    \centering
    \includegraphics[width=\linewidth]{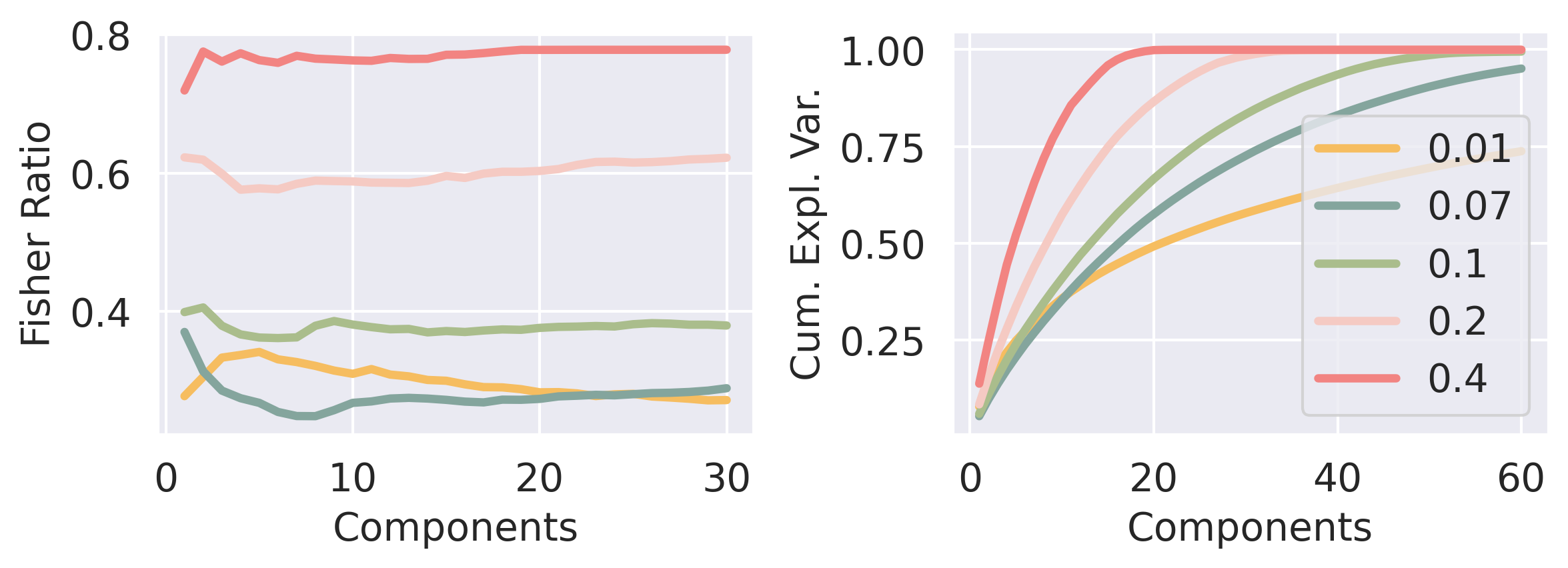}
    \caption{Fisher ratio among clusters and cumulative explained variance by the centroid in the MS COCO dataset. Higher temperatures have a higher Fisher ratio, meaning that the variance inter-cluster is higher than the variance intra-cluster, thus the embeddings are grouped according to the semantics.}
    \label{fig:fisher}
\end{figure}

The temperature $\tau$ controls the softness of the similarity distribution. For small values of $\tau$, the softmax in Eq.~\eqref{eq:infonce} becomes sharper: 
the highest-similarity negatives dominate the denominator, resulting in strong 
instance-level discrimination. This setting favors retrieval tasks, where it is 
important to rank the true positive higher than all other samples 
\cite{chen2020simple, dinu2025effective}. Conversely, larger values of $\tau$ distribute the gradients more evenly across all negatives, encouraging embeddings to form compact groups rather than isolated pairs. This improves group-level structure in the latent space, where samples of the same class are clustered together \cite{dinu2025effective, kukleva2023temperature}. To quantitatively measure this intuituion, we employ the Fisher ratio, which the Fisher ratio, which measures the inter-cluster variance over the intra-cluster variance, defined as: 

\begin{equation}
\label{eq:fisher}
\mathcal{F} = \frac{\operatorname{Tr}(\mathbf{S}_B)}{\operatorname{Tr}(\mathbf{S}_W)},
\end{equation}

\noindent where $\mathbf{S}_B$ is the between-cluster scatter matrix and $\mathbf{S}_W$ is the within-cluster scatter matrix, defined respectively as
\begin{align}
\mathbf{S}_B &= \sum_{c=1}^C N_c (\boldsymbol{\mu}_c - \boldsymbol{\mu})
(\boldsymbol{\mu}_c - \boldsymbol{\mu})^\top, \\
\mathbf{S}_W &= \sum_{c=1}^C \sum_{i=1}^{N_c} 
(\mathbf{z}_{i,c} - \boldsymbol{\mu}_c)(\mathbf{z}_{i,c} - \boldsymbol{\mu}_c)^\top,
\end{align}

\begin{figure*}
    \centering
    \includegraphics[width=0.97\textwidth]{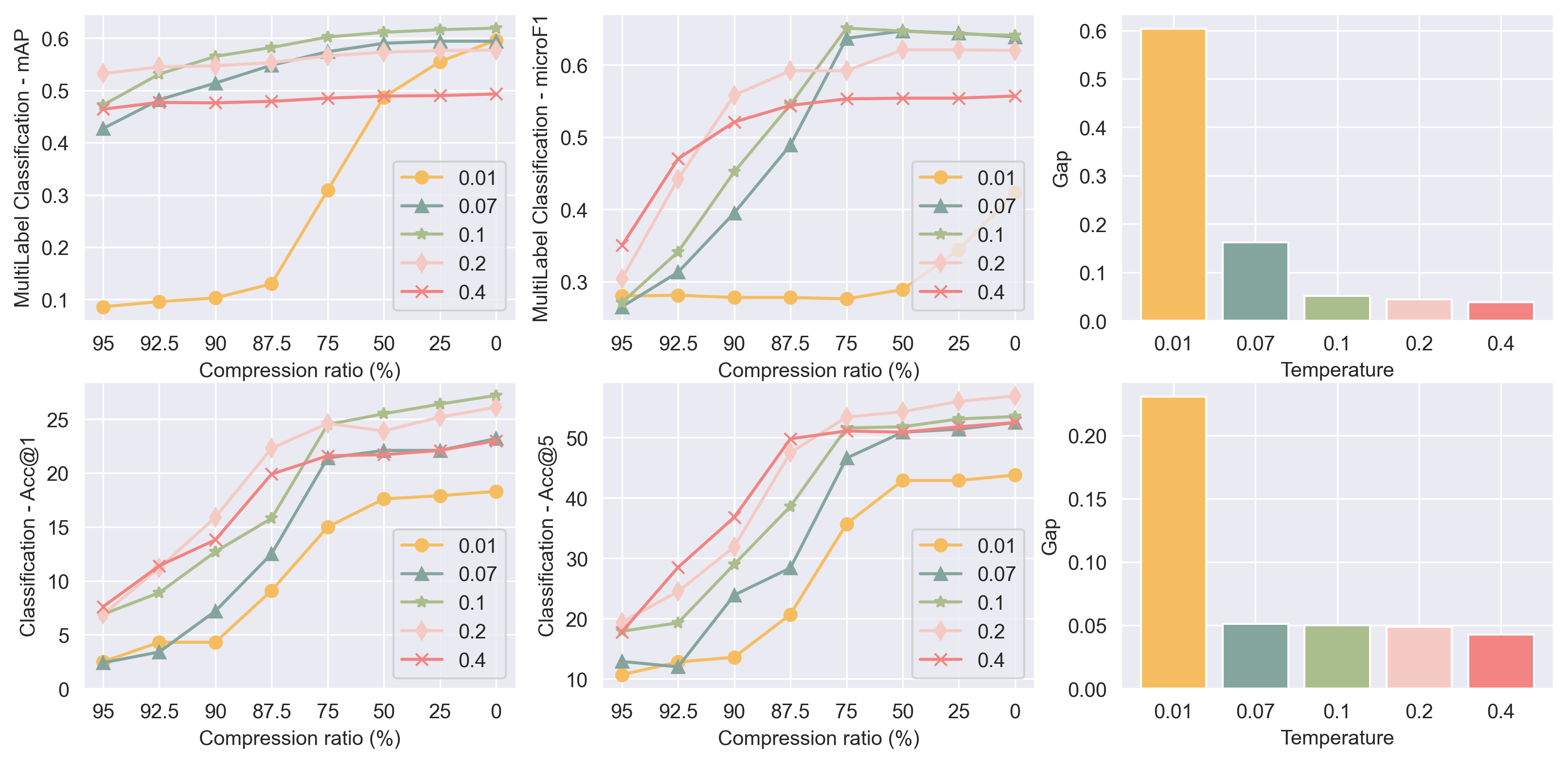}
    \caption{First row: Multilabel classification results on MSCOCO. Second row: Classification results on Flickr30k. In all the experiments, higher temperatures reduce the gap and are more robust to compression, allowing compression without sacrificing performance.}
    \label{fig:mscoco}
\end{figure*}

\noindent with $N_c$ the number of samples in cluster $c$, $\boldsymbol{\mu}_c$ the cluster centroid, and $\boldsymbol{\mu}$ the global mean of all embeddings. Figure~\ref{fig:fisher} shows the plots with the Fisher ratio and the cumulative explained variance. The higher the temperature, the higher the ratio and the fewer components we need to explain the whole variance. This means that, with high temperatures, the difference among clusters is large and the variance inside the same semantic concept cluster is low. On the contrary, with low temperatures, the variance among clusters is very low, badly defining the latent space and consequently hurting group-wise downstream tasks. Such a persistent separation among clusters is commonly named as the modality gap \cite{liang2022mind}, in which embeddings belonging to different modalities are scattered in the space, even when they share the same semantic concept. This phenomenon emerges because contrastive training balances two forces: pulling positive pairs together and spreading apart non-matching pairs.  With low $\tau$, each modality tends to remain in its own region of the latent space, and the gap between these regions stays large. 
As a result, embeddings of semantically matching items do not share the same portion of the space, which harms tasks such as clustering or classification that require coherent group formation. Increasing $\tau$ reduces this effect. By softening the contribution of hard negatives, the embedding spread over the space is more semantically coherent. In practice, higher temperature reduces the modality gap, leading to tighter alignment of matching modalities and a latent space where class centroids become more representative for downstream tasks such as classification, as proven in the unimodal case by \cite{dinu2025effective}.

\subsection{Semantic Compression of Multimodal Embeddings}
\textbf{Centroid representations via gap reduction.} The key intuition of our method is that the modality gap can be reduced by increasing the temperature $\tau$ in the contrastive objective of \eqref{eq:infonce} and this favors embedding compression.
The modality gap can be measured as the Euclidean distance between the centroids of each of the two modalities $m$ and $n$, as in \cite{liang2022mind}:
\begin{equation}
\label{eq:gap}
    \text{Gap}_{m, n} = \| \mathbf{c}^{m} - \mathbf{c}^{n} \|,
\end{equation}
where $\mathbf{c}^{m} = \sum_{t = 1}^N\mathbf{z}_t^m$.
When $\tau$ is sufficiently large, embeddings corresponding to the same semantic concept or pair but originating from different modalities share a common region of the latent space \cite{wu2023understanding, iclr2024two, Eslami2024MitigateTG}, well representing their semantic concepts. Thus, there is no need to store the embeddings from all the modalities, as the same semantics can be stored in a single centroid embedding serving as a compact, modality-agnostic representation of the given semantic concept in the space. This allows a consistent storage saving, as the centroid only requires $1 /\ m$ space.
Formally, for $k=i,j$ semantic concept, we define the centroid representation as
\begin{equation}
\label{eq:centroid}
\boldsymbol{\mu}_k = \frac{1}{M} \sum_{m \in M} \mathbf{z}^m_k.
\end{equation}
Note that the centroid $\boldsymbol{\mu}_k$ is different from $\mathbf{c}^m$ in \eqref{eq:gap}, as $\boldsymbol{\mu}_k$ is the centroid of the semantic concept $k$, while $\mathbf{c}^m$ is the centroid of the modality $m$ cluster.

\textbf{Post-training compression.} While centroids already reduce storage by requiring only one representation per semantic concept, further gains can be obtained through post-training compression. In this paper, we focus on random feature selection (RFS), but any other post-training compression method can be employed as well. In RFS, given a target dimensionality $T < D$, we randomly select $T$ coordinates from $\boldsymbol{\mu}_k$ to form a compressed embedding $\tilde{\boldsymbol{\mu}}_k \in \mathbb{R}^T$:
\begin{equation}
    \tilde{\boldsymbol{\mu}}_{k} \;=\; \mathbf{S}_{\mathcal{I}}\,\boldsymbol{\mu}_{k}, 
    \qquad \mathcal{I}\sim \mathrm{Unif}\!\left\{ I\subseteq[D]\;:\;|I|=T\right\}
\end{equation}
\noindent where $\mathbf{S}_{\mathcal{I}}\in{0,1}^{T\times D}$ is the row-subselection matrix. This method is computationally trivial and preserves performance when embeddings have low intrinsic dimensionality \cite{dinu2025effective}.

\textbf{Compression pipeline.} The proposed compression pipeline consists of three steps:
\begin{enumerate}
    \item Load multimodal pretrained encoders trained with the InfoNCE loss 
    at high temperature to reduce the modality gap. Crucially, we note that very high temperatures are not required, as long as the gap measured as \eqref{eq:gap} remains reasonably lower (empirically checked).
    \item Extract centroid embeddings according to \eqref{eq:centroid}, producing one centroid per semantic class/pair/concept.
    \item Compress centroids via random feature selection and run downstream tasks.
\end{enumerate}
\noindent The resulting representations are compact, modality-agnostic, and 
suitable for downstream classification tasks. Crucially, the gap reduction step ensures that centroids are truly representative of all modalities, 
making post-training compression effective without retraining or modifying 
the encoders.
\section{Experiments}
\label{sec:exp}
\subsection{Setup}
We conduct our evaluation on diverse multimodal benchmarks. The first one is Flickr30k, consisting of 31k images with five captions per image, a dataset widely used in multimodal downstream tasks. Secondly, we consider MSCOCO, which contains more than 330k images annotated with five human-generated captions each, among the most common datasets for multimodal tasks. Different downstream tasks are employed to assess the effect of gap reduction and centroid compression. On both Flickr30k and MSCOCO, we run the classification task in which each image is assigned a label. Since these datasets do not contain label information, we run a large pretrained image classifier to assign ImageNet-1k classes to the test set of both datasets. Then we split the original test set into 80\% and 20\% train-test subsets for the linear classifier and selected the classes with at least 5 samples per class. This results in 552 test samples with 58 classes for Flickr30k and roughly 900 test samples with 183 classes for MSCOCO. We evaluate classification using top-1 and top-5 accuracy.
On MSCOCO, since images contain multiple objects, we also test multi-label classification by assigning multiple labels corresponding to the categories present in the images. We evaluate this task with mean Average Precision (mAP) and microF1 to ensure a balance of all the classes.
\begin{figure}
    \centering
    \includegraphics[width=0.9\linewidth]{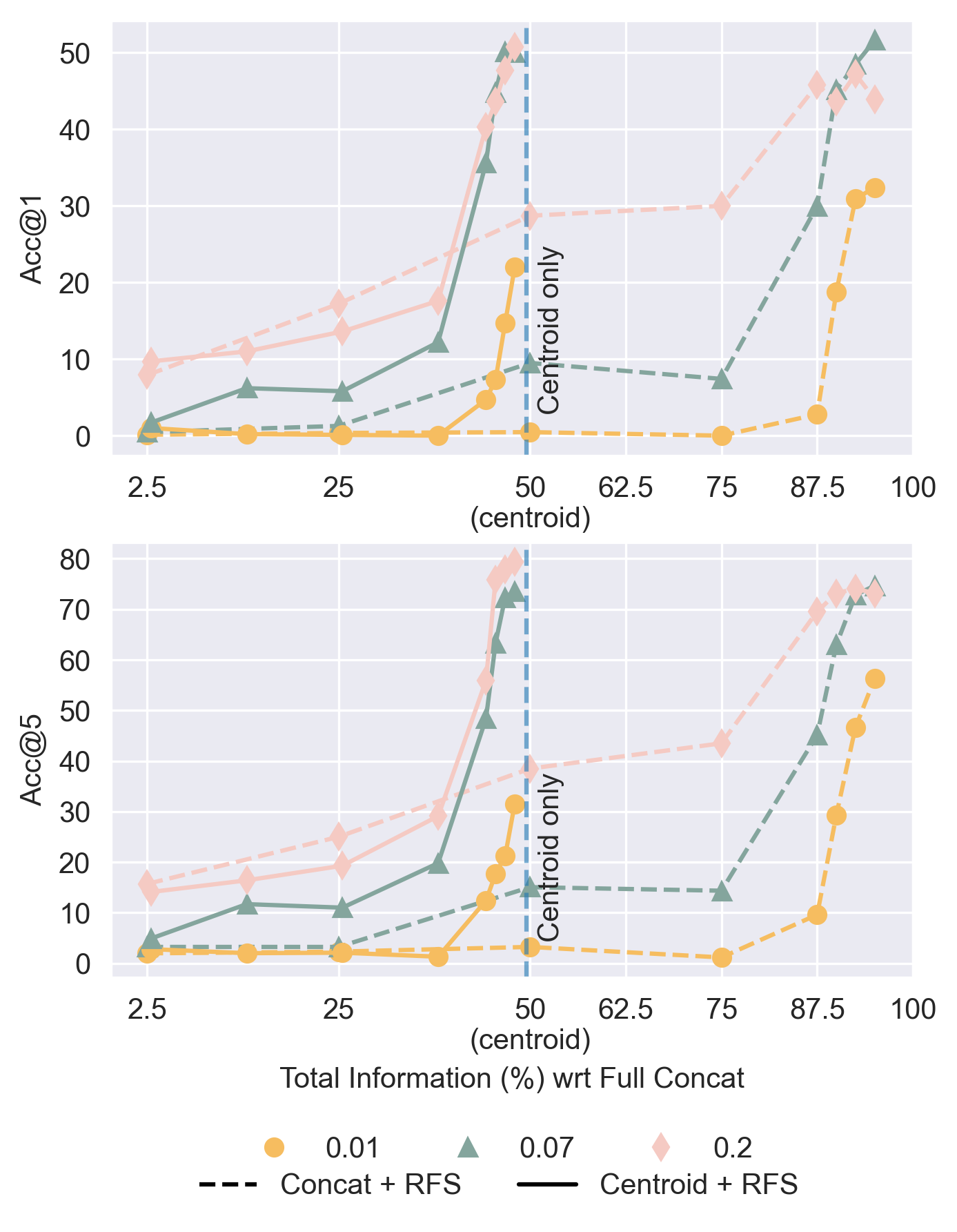}
    \caption{Compression results of proposed method (centroid, solid line) against multimodal embedding concatenation (dashed line) on MSCOCO. Our method achieves comparable or better results with 50\% of total information, enabling effective compression.}
    \label{fig:compression}
\end{figure}

For post-training compression, we adopt random feature selection as our primary technique. After extracting centroids, we randomly select a fixed subset of dimensions to form the compressed representation. This simple strategy is computationally inexpensive and has been shown to perform well when embeddings exhibit redundancy due to low intrinsic dimensionality. We evaluate several compression ratios, ranging from 5\% up to the full dimensionality.
We train the models with OpenCLIP, selecting a ResNet50 backbone (the modality gap and the temperature effect are similar across all the models \cite{iclr2024two} as it depends on the InfoNCE loss) and training for 100 epochs with a learning rate of 0.0001 and different temperatures $\tau$ in \{0.01, 0.07, 0.1, 0.2, 0.4\}. Note that 0.07 is the starting temperature of CLIP, while 0.4 is the one highlighted to be the most oriented to clustering in the unimodal case \cite{dinu2025effective}.
\subsection{Effect of Gap for Compression}
Figure~\ref{fig:mscoco} shows the downstream task results and the gap on MSCOCO and Flickr30k,  first and second row, respectively. For the task of multilabel classification in the first row, we report mAP and microF1. By increasing the temperature from 0.01 to 0.4, the gap decreases, reaching a minimum at 0.4, which impacts compression robustness. Indeed, higher temperatures are more robust to compression, preserving good performance even in the case of 95\% random feature drop, as measured by the mAP. This means that, when the multimodal latent space is semantically organized with a minimum gap, the centroid explains most of the variance and it is extremely representative for a given class. In terms of overall performance, temperatures like 0.1 achieve improved performance when no or little compression is applied, balancing compression robustness and performance at full size, in accordance with the results in the unimodal scenario \cite{dinu2025effective}. Similar considerations can be drawn for the second row on the Flickr30k dataset, in which the model trained with the highest temperature of 0.4 preserves the performance across various compression rates with random feature selection better than all the others. As for the first row of the same figure, a good trade-off between compression robustness and overall performance can be achieved with slightly higher temperatures, which consistently reduce the gap anyway. 

\subsection{Centroid Compression}
Figure~\ref{fig:compression} shows the gain in leveraging the proposed centroid-based compression algorithm over the conventional multimodal embedding concatenation. From those two plots reporting accuracy@1 and accuracy@5 in the classification task on MSCOCO, two results can be drawn. First, involving the proposed centroid compression, we can achieve comparable or even higher scores by involving only 50\% of the information. This means that, without applying any further compression algorithm, our method reduces the size of embeddings by 50\% with no loss in performance. Second, when applying random feature selection (RFS), centroid-based classification better preserves the performance with respect to the full concatenation of the multimodal embeddings. This demonstrates that centroids encapsulate more information and can be further compressed, leading to lower storage requirements and more effective representations.

\section{Conclusion}
In this paper, we investigated the relationship between the modality gap in contrastive multimodal learning and how reducing such a gap enables post-training embedding compression. We showed that when the gap is reduced by training with sufficiently high temperatures, embeddings from different modalities share the same region of the latent space. In this regime, semantic centroids provide faithful, modality-agnostic representations of each class. Building on this observation, we proposed a compression pipeline that first replaces modality-specific embeddings with centroids and then applies lightweight post-training techniques such as random feature selection. This approach does not require retraining and can be applied directly to pretrained models, yielding compact and efficient representations. Our theoretical analysis and empirical validation demonstrate that reducing the modality gap is a key enabler for multimodal embedding compression, opening new directions for scalable and deployment-ready multimodal systems.

\ninept
\balance
\bibliographystyle{IEEEbib}
\bibliography{strings,refs}

\end{document}